\documentclass{article}

\usepackage{PRIMEarxiv}
\usepackage{amsmath}
\usepackage[utf8]{inputenc} 
\usepackage[T1]{fontenc}    
\usepackage{hyperref}       
\usepackage{url}            
\usepackage{booktabs}       
\usepackage{amsfonts}       
\usepackage{nicefrac}       
\usepackage{microtype}      
\usepackage{lipsum}
\usepackage{fancyhdr}       
\usepackage{graphicx}       
\usepackage{cleveref}
\usepackage{subcaption}
\graphicspath{{media/}}     

\title{An analysis of Universal Differential Equations for data-driven discovery of Ordinary Differential Equations}

\author{
  Mattia Silvestri*\\
  University of Bologna\\
  DISI \\
  \texttt{mattia.silvestri4@unibo.it} \\
   \And
  Federico Baldo*\\
  University of Bologna\\
  DISI \\
  \texttt{federico.baldo2@unibo.it} \\
    \And 
  Eleonora Misino*\\
  University of Bologna\\
  DISI \\
  \texttt{eleonora.misino2@unibo.it} \\
    \And 
  Michele Lombardi\\
  University of Bologna\\
  DISI \\
  \texttt{michele.lombardi2@unibo.it} \\
}

\newcommand{\printfnsymbol}[1]{%
  \textsuperscript{\@fnsymbol{#1}}%
}
\makeatother

\newcommand{\norm}[1]{\left\lVert#1\right\rVert}

\newcommand{\uodeacronym}{UDE}
\newcommand{\pairmethod}{\textsc{mini-batch}}
\newcommand{\unrollmethod}{\textsc{full-batch}}

\begin{document}
\maketitle
\def\thefootnote{*}\footnotetext{Equal contribution.}\def\thefootnote{\arabic{footnote}}

\begin{abstract}
In the last decade, the scientific community has devolved its attention to the deployment of data-driven approaches in scientific research to provide accurate and reliable analysis of a plethora of phenomena. 
Most notably, Physics-informed Neural Networks and, more recently, Universal Differential Equations (UDEs) proved to be effective both in system integration and identification. However, there is a lack of an in-depth analysis of the proposed techniques. In this work, we make a contribution by testing the UDE framework in the context of Ordinary Differential Equations (ODEs) discovery. In our analysis, performed on two case studies, we highlight some of the issues arising when combining data-driven approaches and numerical solvers, and we investigate the importance of the data collection process. We believe that our analysis represents a significant contribution in investigating the capabilities and limitations of Physics-informed Machine Learning frameworks.
\end{abstract}


\section{Introduction} 


Physics-informed Machine Learning has gained high attention in the last few years \cite{chen2018neural,lai2021structural,bradley2021two,raissi2019physics,zhang2019quantifying,rackauckas2020universal,O_Leary_2022}, enabling the integration of physics knowledge into machine learning models.
Purely data-driven methods, like Deep Neural Networks (DNNs), have huge representational power and can deal with noisy high dimensional raw data; however, they may learn observational biases, leading to physically inconsistent predictions and poor generalization performance. 
On the other hand, despite the relentless progress in the field, solving real-world partial differential equations (PDEs) using traditional analytical or computational approaches requires complex formulations and prohibitive costs. 
A lot of effort has been devoted to bridging DNNs with differential equations in end-to-end trainable frameworks. 
However, less attention has been paid to analyze the advantages and limitations of the proposed approaches.\\\\
We view the lack of an in-depth analysis of physics-informed techniques as a major issue. We make a contribution in this area by performing an analysis on the Universal Differential Equation (\uodeacronym) \cite{rackauckas2020universal} framework in the context of data-driven discovery of ODEs.  We focus on \uodeacronym{} since its general formulation allows to express other existing frameworks.
In particular, we focus on: 1) \emph{evaluating two training approaches in terms of accuracy and efficiency}; 2) \emph{testing the effect of the numerical solver accuracy in the parameters approximation}, 3) \emph{analyzing the impact of the data collection process regarding the approximation accuracy}, and in 4) \emph{exploring the effectiveness of  \uodeacronym{} in  reconstructing a functional dependence between a set of observables and the unknown parameters}.\\\\
%
The paper is structured as follows. In Section \ref{sec:related}, we provide an overview of the existing work in physics-informed machine learning and system identification. We briefly introduce the \uodeacronym{}  framework in Section \ref{sec:background}, and we describe our research questions in Section \ref{sec:method}. In Section \ref{sec:results}, we present the experiments and report the results. Finally, in Section \ref{sec:conclusions}, we draw some conclusions and discuss future directions.


\section{Related work}
\label{sec:related}

In this section, we briefly present some of the most promising trends in Physics-informed Machine Learning. For an exaustive literature overview, we refer the reader to \cite{karniadakis2021physics}.

\subsection{Physics-informed loss function.}
The most straightforward way to enforce constraints in Neural Networks is via an additional term in the loss function. In \cite{daw2017physics} the authors propose Physics-guided Neural Network, a framework that exploits physics-based loss functions to increase deep learning models' performance and ensure physical consistency of their predictions. Similarly, the work of Chen et al. \cite{jia2019physics} generalizes Recurrent Neural Networks adding a regularization loss term that captures the variation of energy balance over time in the context of lake temperature simulation. Work of \cite{beucler2021enforcing} proposes to enforce physics constraints in Neural Networks by introducing a penalization term in the loss function defined as the mean squared residuals of the constraints.

\subsection{Physics-informed neural architectures.}
Recent works focus on designing deep learning frameworks that integrate physics knowledge into the architecture of deep learning models \cite{chen2018neural,lai2021structural,bradley2021two,raissi2019physics,zhang2019quantifying,rackauckas2020universal,O_Leary_2022}. Neural Ordinary Differential Equations (Neural ODEs) \cite{chen2018neural} bridge neural networks with differential equations by defining an end-to-end trainable framework. In a Neural ODE, the derivative of the hidden state is parameterized by a neural network, and the resulting differential equation is numerically solved through an ODE solver, treated as a black-box. Neural ODEs have proven their capacity in time-series modeling, supervised learning, and density estimation. Moreover, recent works adopt Neural ODEs for system identification by learning the discrepancy between the prior knowledge of the physical system and the actual dynamics \cite{lai2021structural} or by relying on a two-stage approach to identify unknown parameters of differential equations \cite{bradley2021two}. Recently, O'Leary et al. \cite{O_Leary_2022} propose a framework that learns hidden physics and dynamical models of stochastic systems. Their approach is based on Neural ODEs, moment-matching, and mini-batch gradient descent to approximate the unknown hidden physics.
Another approach is represented by the Physics-informed Neural Network (PINN) framework \cite{raissi2019physics} which approximates the hidden state of a physical system through a neural network. The authors show how to use PINNs both to solve a PDE given the model parameters and to discover the model parameters from data. 
Zhang et al. \cite{zhang2019quantifying} further extend PINNs by accounting for the uncertainty quantification of the solution. In particular, the authors focus on the \textit{parametric} and \textit{approximation uncertainty}. 
Universal Differential Equations (UDEs) \cite{rackauckas2020universal} represent a generalization of Neural ODE where part of a differential equation is described by a universal approximator, such as a neural network. The formulation is general enough to allow the modeling of time-delayed, stochastic, and partial differential equations. Compared to PINNs, this formalism is more suitable to integrate recent and advanced numerical solvers, providing the basis for a library that supports a wide range of scientific applications.
%
%
\subsection{System identification.} Research towards the automated dynamical system discovery from data is not new \cite{DBLP:journals/compsys/CrutchfieldM87}. The seminal works on system identification through genetic algorithms \cite{DBLP:journals/pnas/BongardL07,doi:10.1126/science.1165893} introduce symbolic regression as a method to discover nonlinear differential equations. However, symbolic regression is limited in its scalability.  Brunton and Lipson  \cite{brunton2016discovering} propose a sparse regression-based method for identifying ordinary differential equations, while Rudy et al. \cite{doi:10.1126/sciadv.1602614} and Schaeffer \cite{schaeffer2017} apply sparse regression to PDEs discovering. 
Recent works \cite{lu2020extracting,lai2021structural,bradley2021two} focus on applying physics-informed neural architectures to tackle the system discovery problem. Lu et al. \cite{lu2020extracting} propose a physics-informed variational autoencoder to learn unknown parameters of dynamical systems governed by partial differential equations. The work of Lai et al. \cite{lai2021structural} relies on Neural ODE for structural-system identification by learning the discrepancy with respect to the true dynamics, while Bradley at al. \cite{bradley2021two} propose a two-stage approach to identify unknown parameters of differential equations employing Neural ODE. 
\section{Universal Differential Equations}
\label{sec:background}
The Universal Differential Equation (\uodeacronym{}) \cite{rackauckas2020universal} formulation relies on embedded universal approximators to 
model forced stochastic delay PDEs in the form:

    \begin{equation}
        \mathcal{N} \left[ u(t), u(\alpha(t)), \text{W}(t), U_{\theta}(u, \beta(t)) \right] = 0
        \label{eq:\uodeacronym{}}
    \end{equation}
where $u(t)$ is the system state at time $t$,  $\alpha(t)$ is a delay function, and $\text{W}(t)$ is the Wiener process. $\mathcal{N} \left[ \cdot \right]$ is a nonlinear operator and $U_{\theta}(\cdot)$ is a universal approximator parameterized by $\theta$. 
The \uodeacronym{} framework is general enough to express other frameworks that combine physics knowledge and deep learning models. For example, by considering a one-dimensional \uodeacronym{} defined by a neural network, namely $u'= U_\theta (u(t), t)$, we retrieve the Neural Ordinary Differential Equation framework \cite{chen2018neural,NEURIPS2019_42a6845a,NEURIPS2020_4a5876b4}.  

\uodeacronym s are trained by minimizing a cost function $C_\theta$ defined on the current solution $u_\theta(t)$ with respect to the parameters $\theta$. The cost function is usually computed on discrete data points $(t_i,y_i)$ which represent a set of measurements of the system state, and the optimization can be achieved via gradient-based methods like ADAM \cite{kingma2017adam} or Stochastic Gradient Descent (SGD) \cite{robbins1951stochastic}.




\section{\uodeacronym{} for data-driven discovery of ODEs}
\label{sec:method}
In this section, we present the \uodeacronym{} formulation we adopt, and we describe four research questions aimed at performing an in-depth analysis of the UDEs framework in solving data-driven discovery of ODEs. 
\subsection{Formulation} We restrict our analysis to dynamical systems described by ODEs with no stochasticity or time delay. The corresponding \uodeacronym{} formulation is:
\begin{equation}
    u^{\prime}=f(u(t),t,U_{\theta}(u(t),t)).
        \label{eq:universalode}
\end{equation}
where $f(\cdot)$ is the known dynamics of the system, and $U_\theta(\cdot,\cdot)$ is the universal approximator for the unknown parameters.
As cost function, we adopt the Mean Squared Error (MSE) between the current approximate solution $u_\theta(t)$ and the true measurement $y(t)$, formally:
    \begin{equation}
        C_{\theta}=\sum_{i} \norm{u_{\theta}(t_i)-y(t_i)}_{2}^2.
    \label{eq:uodecostfunction}
    \end{equation}
We consider discrete time models, where the differential equation in \eqref{eq:universalode} can be solved via numerical techniques. Among the available solvers, we rely on the Euler method, which is fully differentiable and allows for gradient-based optimization. Moreover, the limited accuracy of this first-order method enlightens the effects of the integration technique on the unknown parameter approximation.
Our analysis starts from a simplified setting, in which we assume that the unknown parameters are fixed. Therefore, the universal approximator in \Cref{eq:universalode} reduces to a set of learnable variables, leading to:
\begin{equation}
    u^{\prime}=f(u(t),t,U_{\theta}).
        \label{eq:uode_fixed_param}
\end{equation}
%

\subsection{Training Procedure}

Given a set of state measurements $y$ in the discrete interval $[t_0,t_n]$, we consider two approaches to learn Equation \eqref{eq:uode_fixed_param}, which we analyze in terms of \emph{accuracy} and \emph{efficiency}. The first approach, mentioned by \cite{raissi2019physics} and named here \unrollmethod{}, involves 1) applying the Euler method on the whole temporal series with $y(t_0)$ as the initial condition, 2) computing the cost function $C_{\theta}$, and 3) optimizing the parameters $\theta$ via full-batch gradient-based methods.
An alternative approach, named \pairmethod{}, consists of splitting the dataset into pairs of consecutive measurements $(y(t_i),y(t_{i+1}))$, 
and considering each pair as a single initial value problem. Then, by applying the Euler method on the single pair, we can perform a mini-batch training procedure, which helps in mitigating the gradient vanishing problem \cite{glorot2010understanding}.
Conversely to the \unrollmethod{} approach, which requires data to be ordered and uniform in observations, the \pairmethod{} method has less strict requirements and can be applied also to partially ordered datasets.

\subsubsection{Solver Accuracy}
In the \uodeacronym{} framework, the model is trained to correctly predict the system evolution by learning an approximation of the unknown parameters that 
minimizes the cost function $C_{\theta}$. 
The formulation relies on the integration method to approximate the system state $u(t)$. However, the numerical solver may introduce approximation errors that affect the whole learning procedure.  Here, we want to investigate the \emph{impact of the solver accuracy on the unknown parameters approximation}.
Since the Euler method is a first-order method, its error depends on the number of iterations per time step used to estimate the value of the integral, and, thus,  we can perform our analysis with direct control on the trade-off between execution time and solver accuracy.


\subsection{Functional Dependence}
By relying on the universal approximator in \Cref{eq:universalode}, the \uodeacronym{} framework is able to learn not only fixed values for the unknown parameters, but also functional relationships between them and the observable variables.  Thus, we add a level of complexity to our analysis by considering the system parameters as functions of observable variables, and we \emph{evaluate the \uodeacronym{} accuracy in approximating the unknown functional dependence.}


\subsubsection{Data Sampling}
Since \uodeacronym{} framework is a data-driven approach, it is important to investigate the effectiveness of the \uodeacronym{} framework under different data samplings. In particular, \emph{can we use the known dynamics of the system under analysis to design the data collection process in order to increase the approximation accuracy?}

\section{Empirical Analysis}
\label{sec:results}
Here, we report the results of our analysis performed on two case studies: 1)  \textit{RC circuit}, i.e.,  estimating the final voltage in a first-order resistor-capacitor circuit; 2) \textit{Predictive Epidemiology}, i.e., predicting the number of infected people during a pandemic.
We start by describing the two case studies; then, we illustrate the evaluation procedure and the experimental setup. Finally, we present the experiments focused on the research questions highlighted in Section \ref{sec:method}, and we report the corresponding results.
%
\subsection{RC Circuit}
We consider a first-order RC circuit with a constant voltage generator. The state evolution of the system is described by
\begin{equation}
    \frac{dV_{C}(t)}{dt}=\frac{1}{\tau}(V_s - V_C(t))
 \label{eq:rcderivative}
\end{equation}
where $V_C(t)$ is the capacitor voltage at time $t$, $V_s$ is the voltage provided by the generator, and $\tau$ is the time constant which defines the circuit response.

We use the \uodeacronym{} formulation to approximate  $\tau$ and $V_s$ by writing Equation \eqref{eq:rcderivative} as
    \begin{equation}
        u^{\prime}=\frac{1}{U_{\theta_1}(t)}(U_{\theta_2}(t) - u(t))
     \label{eq:uderc}
    \end{equation}
where $u_t$ is a short notation for $V_C(t)$, $U_{\theta_1}(t)$ and $U_{\theta_2}(t)$ are the neural networks approximating $\tau$ and $V_s$ respectively.
The cost function is defined as 
    \begin{equation}
        C_{\theta_1,\theta_2} = \sum_{i}(u_{\theta_1,\theta_2}(t_i)-y_i)^2 
    \end{equation}
where $u_{\theta_1,\theta_2}(t_i)$ and $y_i$ are the current solution and the discrete-time measurements of the capacitor voltage at time $t_i$, respectively.

\subsection{Predictive Epidemiology}
Among the different compartmental models used to describe epidemics, we consider the well-known Susceptible-Infected-Recovered (SIR) model, where the disease spreads through the interaction between susceptible and infected populations.
The dynamics of a SIR model is described by the following set of differential equations:
\begin{align}
    \begin{split}
            \frac{dS}{dt} &= - \beta~\frac{S\cdot I}{N}, \\ 
        \frac{dI}{dt} &= \beta~\frac{S\cdot I}{N} - \gamma~I, \\ 
        \frac{dR}{dt} &= \gamma~I ,
    \end{split}
    \label{eq:sir}
\end{align}
where $S,I$, and $R$ refer to the number of susceptible, infected, and recovered individuals in the population. The population is fixed, so $N = S+I+R$. The parameter $\gamma \in [0,1]$ depends on the average recovery time of an infected subject, while $\beta \in [0,1]$ is the number of contacts needed per time steps to have a new infected in the susceptible population. $\beta$ determines the spreading coefficient of the epidemic and is strongly affected by different environmental factors (e.g.,  temperature, population density, contact rate, etc.). The introduction of public health measures that directly intervene on these environmental factors allows to contain the epidemic spreading.

We rely on the \uodeacronym{} framework to i) perform system identification on a simulated SIR model, and ii) estimate the impact of \textit{Non-Pharmaceutical Interventions} (NPIs) on the epidemic spreading.
We define the state of the system at time $t$ as $\mathbf{u}_t = (S_t,I_t,R_t)$ and we formulate the Equations in \eqref{eq:sir} as



%
\begin{equation}
    \mathbf{u}' = f(\mathbf{u}_t, t, U_{\theta}(\mathbf{u}_t, t, X_t))
\end{equation}

where 
$X_t$ is the set of NPIs applied at time $t$. We assume $\gamma$ to be fixed and known, and we approximate the  SIR model parameter $\beta$ with a neural network $U_{\theta}(\mathbf{u}_t, t, X_t)$. The cost function for this case study is defined as 
    \begin{equation}
        C_{\theta} = \sum_{i}(u_{\theta}(t_i)-\hat{y}_i)^2 
    \end{equation}
where $u_{\theta}(t_i)$ and $y_i$ are the current solution and the discrete-time measurements of the system state at time $t_i$, respectively.


\subsection{Evaluation and experimental setup.} We evaluate the model accuracy by relying on two metrics: the \emph{Absolute Error} (AE), to evaluate the estimation of the parameters, and the \emph{Root Mean Squared Error} (RMSE), to study the approximation of the state of the dynamic systems. 
For each experiment, we perform $100$ trials, normalize the results, and report mean and standard deviation.
All the experiments are run on a Ubuntu virtual environment equipped with 2 Tesla V100S, both with a VRAM of 32 GB. We work in a \texttt{Python 3.8.10} environment, and the neural models are implemented in \texttt{TensorFlow 2.9.0}. The source code is available at \url{https://github.com/ai-research-disi/ode-discovery-with-ude}.
%

\subsection{Training Procedure}
\label{sec:training_procedure}
We compare \unrollmethod{} and \pairmethod{} methods to assess which is the most accurate and efficient. 
We rely on high-precision simulation to generate data for both case studies. 
For the RC circuit, we set $V_c(0) = 0$, and we sample $100$ values of $V_s$ and $\tau$ in the range $\left[ 5, 10 \right]$ and $\left[ 2, 6 \right]$, respectively. Then, we generate data by relying on the analytical solution of Equation \ref{eq:rcderivative}. From each of the resulting curves, we sample $10$ data points $(V_c(t),t)$ equally spaced in the temporal interval $\left[ 0, 5 \tau \right]$. 
Concerning the epidemic case study, the data generation process relies on a highly accurate Euler integration with $10.000$ iterations per time step. 
We use the same initial condition across all instances, namely $99\%$ of susceptible and  $1\%$ of infected on the entire population, and we assume $\gamma$ to be equal to $0.1$, meaning that the recovery time of infected individuals is on average $10$ days. We create $100$ epidemic curves, each of them determined by the sampled value of  $\beta$ in the interval $\left[ 0.2, 0.4\right]$. 
The resulting curves contain daily data points in the temporal interval from day $0$ to day $100$ of the outbreak evolution.

We evaluate the accuracy of \uodeacronym{} in approximating the unknown parameters and the system state, and we keep track of the total computation time required to reach convergence. 
We believe it is relevant to specify that the \pairmethod{} has an advantage compared to the \unrollmethod{}. The evaluation of the latter involves predicting the the whole state evolution given only the initial one $u_0$; whereas, the first approach reconstructs the state evolution if provided with intermediate values. Thus, to have a fair comparison, the predictions of the \pairmethod{} method are fed back to the model to forecast the entire temporal series given only $u_0$.
As shown in \Cref{table:batch_rc_det}, for the RC circuit case study, both \unrollmethod{} and \pairmethod{} approximate quite accurately $V_s$ and $V_c(t)$, whereas the approximation of $\tau$ has a non-negligible error. However, \unrollmethod{} requires almost $3$ times the computational time to converge. In the SIR use case (\Cref{table:batch_sir_det}), the two training procedures achieve very similar accuracies,  but \unrollmethod{} is more than $8$ times computationally expensive.
Since both the methods have very similar estimation accuracy, we can conclude that \textit{\pairmethod{} is a more efficient method to train the \uodeacronym{} compared to \unrollmethod{}}. Thus, we rely on the \pairmethod{} method in the remaining experiments.
\begin{table*}
\centering
\caption{Comparison between \pairmethod{} and \unrollmethod{} methods in RC circuit use case. We report the AE of $V_s$ and $\tau$ approximation, RMSE for $V_c(t)$ prediction, and computational time in seconds.}
\label{table:batch_rc_det}
\vspace{2pt}
\addtolength{\tabcolsep}{4pt}
\small
\begin{tabular}{lcccc}
\toprule
{} & {$V_s$}  & {$\tau$} & {$V_c(t)$} & {Time}\\
\midrule
     \pairmethod{}   & $0.027 \pm 0.013$  & $0.163 \pm 0.101$  & $0.021 \pm 0.010$  & $9.21 \pm 39.49$  \\
    \unrollmethod{} &   $0.018 \pm 0.021 $  & $0.200 \pm 0.081 $  & $0.014 \pm 0.020$  & $26.19 \pm 5.69$\\
\bottomrule
\end{tabular}
\label{tab:exp1det}
\end{table*} 
\begin{table*}[htp]
\centering
\caption{Comparison between \pairmethod{} and \unrollmethod{} methods in epidemic use case. We report the AE of $\beta$ approximation, RMSE for $SIR(t)$ prediction, and computational time in seconds.}
\label{table:batch_sir_det}
\vspace{2pt}
\addtolength{\tabcolsep}{4pt}
\small
\begin{tabular}{lcccc}
\toprule
{} & {$\beta$} & {$SIR(t)$} & {Time}\\
\midrule
     \pairmethod{}   &  
    $0.0030 \pm 0.0019$  & $0.017 \pm 0.0046$  & $1.28 \pm 0.23$ \\
    \unrollmethod{} &  
    $0.0065 \pm 0.0053$ & $0.019 \pm 0.0079$ & $10.23 \pm 2.50$ \\
\bottomrule
\end{tabular}
\label{tab:batch_rc_det}
\end{table*} 
%
\subsection{Solver Accuracy}
\label{sec:solveraccuracy}
\begin{figure}[htp]
   \centering
   \includegraphics[width=.6\linewidth]
   {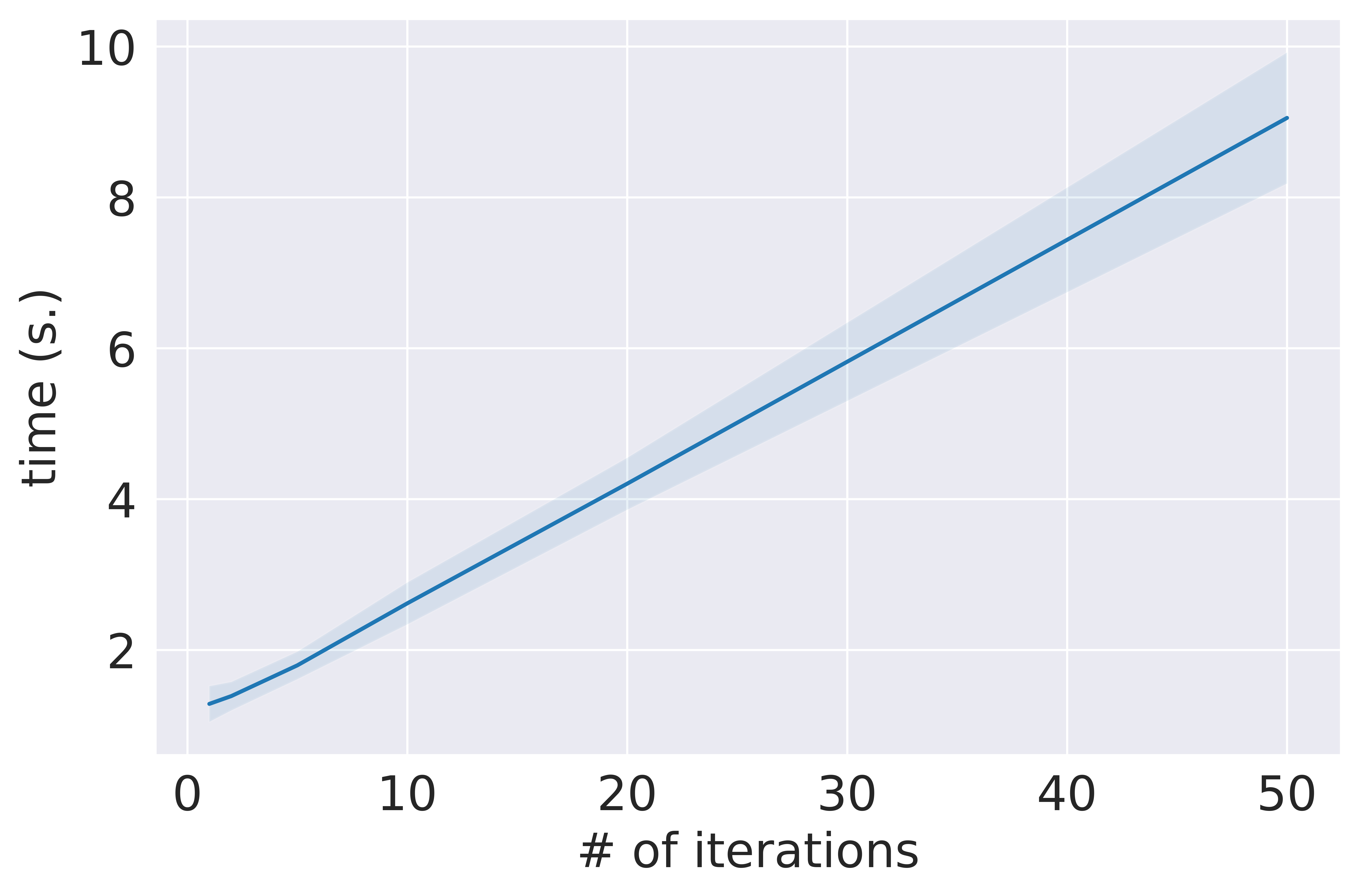}
   \caption{\uodeacronym{} training time as a function of the number of iterations per time step of the Euler method.}
   \label{fig:exp2-time}
 \end{figure}

In the context of ODE discovery, we are interested in approximating the unknown system parameters. Despite an overall accurate estimation of the system state, the results of the previous analysis show that \uodeacronym{} framework does not reach high accuracy in approximating the system parameters. 
%
The model inaccuracy might be caused by the approximation error introduced by the integration method. 
Thus, to investigate the \emph{impact of the solver accuracy on the unknown parameters approximation}, we test different levels of solver accuracy by increasing the number of iterations between time steps in the integration process.
A higher number of iterations per time step of the Euler method should lead to more accurate solutions of the ODE; however, this comes also at the cost of a higher computational time as shown in Figure \ref{fig:exp2-time}.

\begin{figure}[htp]
 \begin{subfigure}{.45\textwidth}
   \centering
   \includegraphics[width=1\linewidth]
   {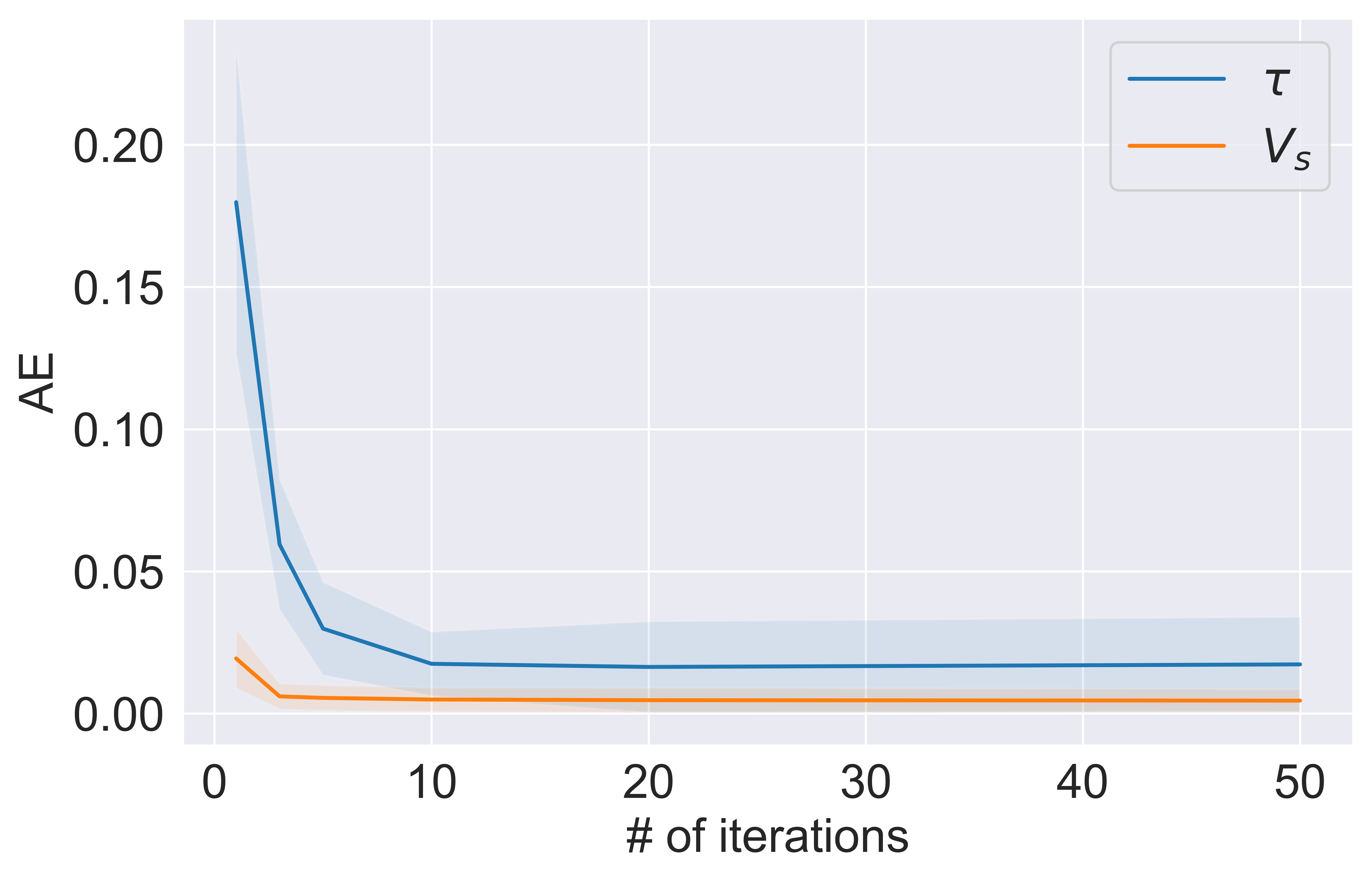}
   \caption{$\tau$ and $V_s$}
   \label{fig:exp2-sfig1}
 \end{subfigure}
 \begin{subfigure}{.45\textwidth}
   \centering
   \includegraphics[width=1\linewidth]
   {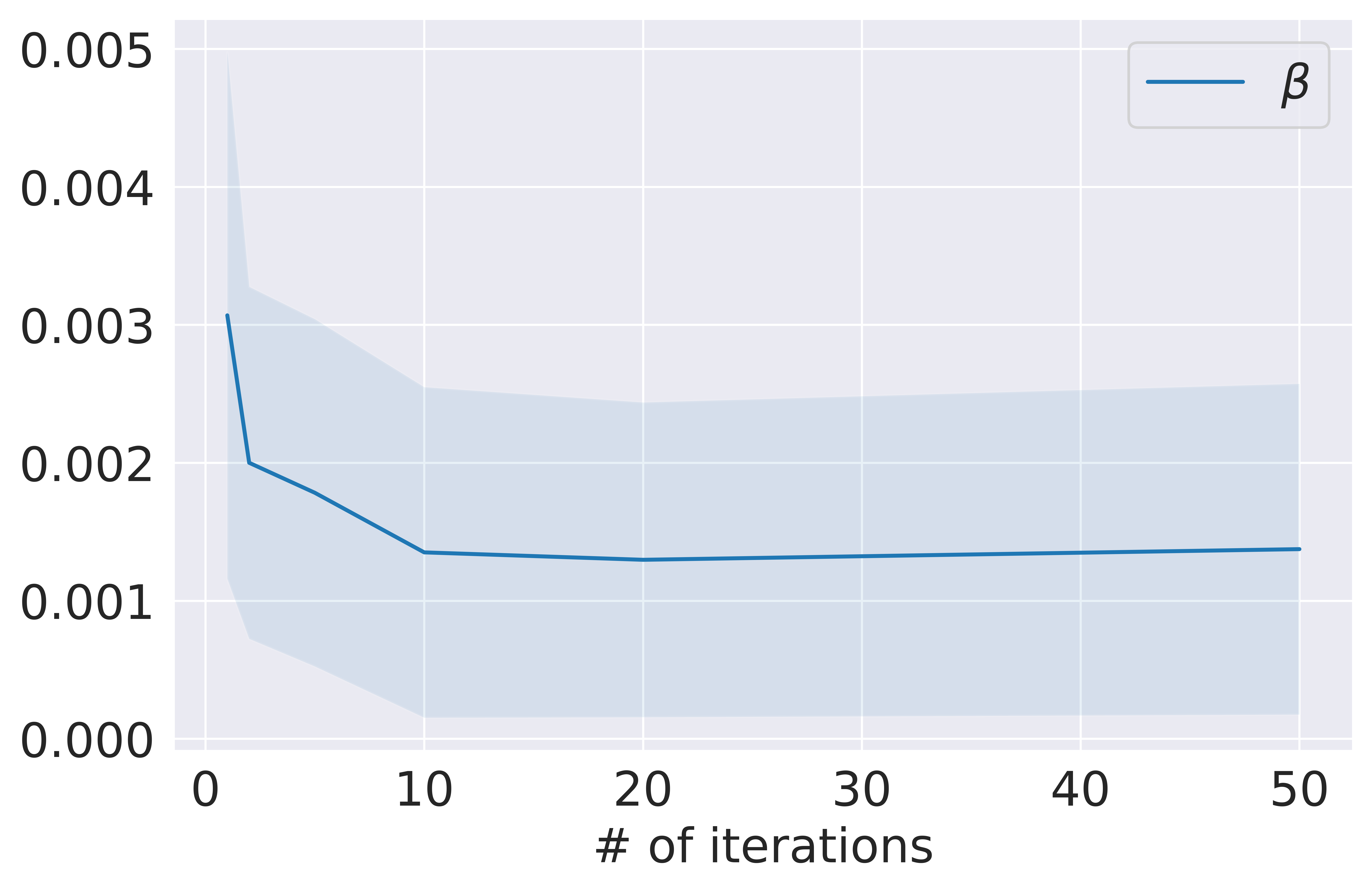} 
   \caption{$\beta$}
   \label{fig:exp2-sfig2}
\end{subfigure}

\caption{Average and standard deviation of the AE as a function of the number iterations per time step of the Euler method.}
\label{fig:exp2}

\end{figure}

In this experiment, we use the same data generated for \emph{Training procedure} experiment.
In \Cref{fig:exp2}, we report the approximation error of the \uodeacronym{} framework when applying the Euler method with an increasing number of steps. As expected, in both use cases, by increasing the precision of the Euler method, the ODE parameters estimation becomes more accurate, until reaching a plateau after $10$ iterations per time step. 


\subsection{Functional Dependence and Data Sampling}

In a real-world scenario, the dynamical systems that we are analyzing often depend on a set of external variables, or \emph{observables}, that influence the behaviour of the system.
These elements can be environmental conditions or control variables which affect the evolution of the system state. 
We study the \uodeacronym{} framework in presence of observables, assuming two kinds of relationship between the independent and dependent variable, namely, a \emph{linear} and a \emph{non-linear} dependence.

\subsubsection{Linear Dependence}
For the RC circuit, we consider a simple and controlled setup where $\tau$ is a linear function of a continuous input variable $x$ changing over time, namely $\tau(x) = ax$, where $a$ and $x$ are scalar values. Conversely to the previous experiments, we assume $V_s$ to be known and equal to $1$; we perform this design choice to focus our analysis on the approximation accuracy of the linear relationship solely.
Since the value of $\tau$ changes over time, we can not rely on the analytic solution of Equation \eqref{eq:rcderivative} to generate data. Thus, we generate samples from one timestep to the successive one by running a high-resolution integration method, namely the Euler method with $10,000$ iterations per time step.
In the generation process, the linear coefficient $a$ is randomly sampled from a uniform probability distribution in the interval $\left[ 2, 6 \right]$, and the observable $x$ is initialized to $1$, and updated at each time step as follows:
$$x(t) = x(t-1) + \epsilon, \quad \text{with} \quad \epsilon \sim \mathcal{U}_{[0,1]}.$$
This procedure allows to have reasonable variations of $\tau$ to prevent physically implausible data.
During the learning process, as a consequence of the results obtained in the \emph{Solver Accuracy} experiment (\Cref{sec:solveraccuracy}),  we use $10$ iterations per time step in the Euler method as a trade-off between numerical error and computational efficiency.

In this experiment, we are interested in \emph{evaluating
the \uodeacronym{} accuracy in approximating the unknown linear dependence.}
The resulting absolute error of the approximation of the linear coefficient $a$ is $0.24 \pm 0.27$. 
Since the \uodeacronym{} is a data-driven approach, the estimation error may be due to the data quality. Since we simulate the RC circuit using a highly accurate integration method resolution, we can assume that data points are not affected by noise. 
However, the sampling procedure may have a relevant impact on the learning process. The time constant $\tau$ determines how quickly $V_c(t)$ reaches the generator voltage $V_s$, and its impact is less evident in the latest stage of the charging curve. Thus, sampling data in different time intervals may affect the functional dependence approximation. To investigate \emph{how data sampling affects the linear coefficient estimation}, we generate $10$ data points in different temporal regions of the charging curve. We consider intervals of the form $[0, EOH]$, where $EOH \in (0,5\tau]$ refers to the end-of-horizon of the measurements; since $\tau$ changes over time, we consider the maximum as a reference value to compute the $EOH$. 
\begin{figure}[htp]
   \centering
   \includegraphics[width=1\linewidth]
   {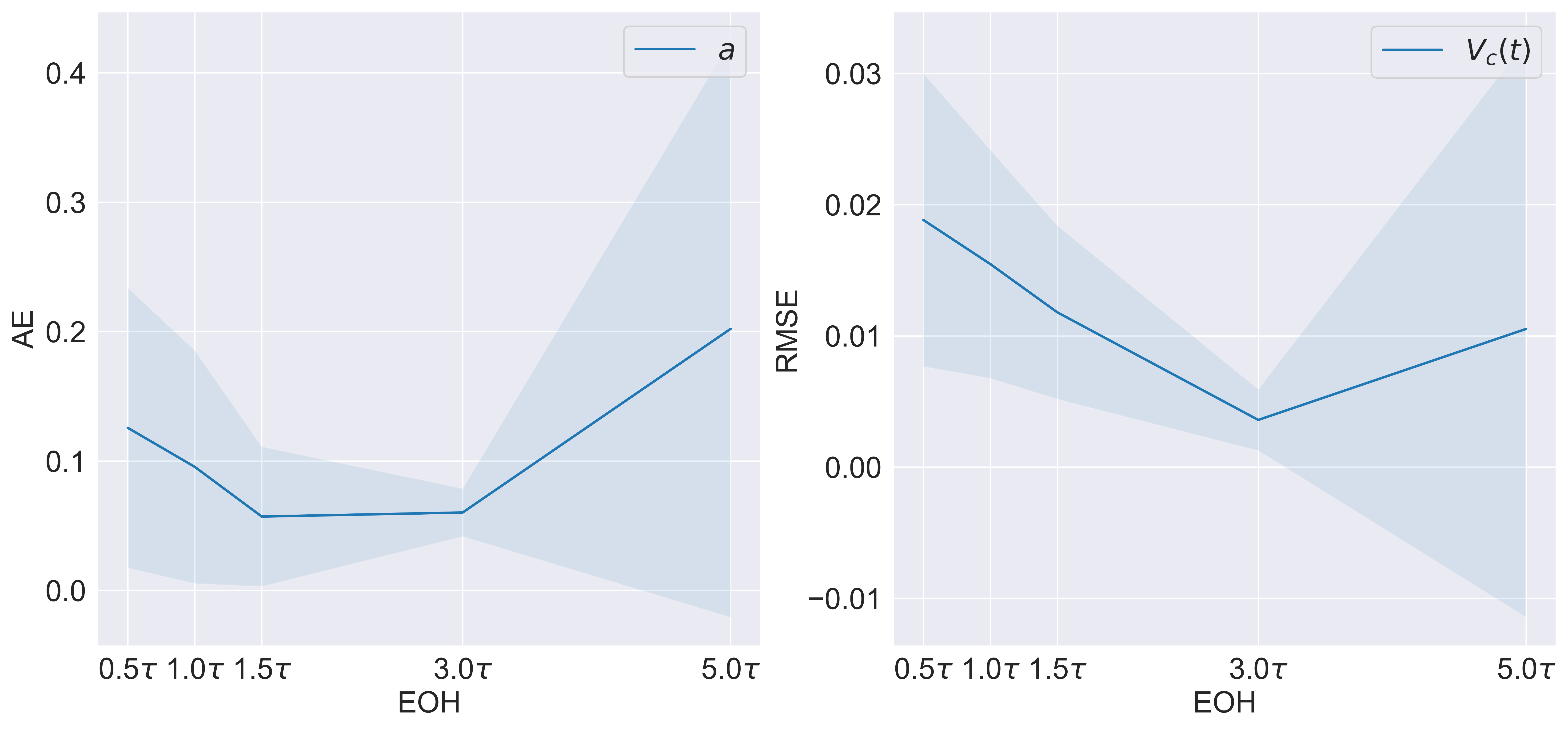}
 \caption{Linear coefficients and predictions error and a function of the EOH.} 
 \label{fig:exp5}
 \end{figure}

As shown in \Cref{fig:exp5}, the linear model approximation is more accurate if the data points are sampled in an interval with $EOH \in [1.5 \tau, 3\tau]$, where $V_c(t)$ approximately reaches respectively the $77\%$ and $95\%$ of $V_s$. With higher values of $EOH$, the sampled data points are closer to the regime value $V_s$, and the impact of $\tau$ is less relevant in the system state evolution. Thus, the learning model can achieve high prediction accuracy of $V_c(t)$ without correctly learning the functional dependence.

\subsubsection{Non-Linear Dependence} Here, we test the \uodeacronym{} framework under the assumption of a non-linear dependence between the observable and the $\beta$ parameter of the epidemic model.
The observable is a set of Non-Pharmaceutical Interventions (NPIs), which affects the virus spreading at each time step. 
To generate the epidemic data, we define the following time series representing the variation at time $t$ of the inherent infection rate of the disease, $\hat{\beta}$, under the effect of two different NPIs per time instance:
\begin{equation}
    \beta(t,\mathbf{x}^t, \mathbf{e},\hat{\beta})= \hat{\beta} \cdot e_1^{x^t_1} \cdot e_2^{x^t_2}
    \label{eq:beta_series}
\end{equation}
where $\mathbf{x}^t \in \{0, 1\}^{2}$ is the binary vector indicating whether the corresponding NPIs are active at time $t$. The vector  $\mathbf{e} \in  \left[0, 1 \right]^2$ represents the effects of the two NPIs in reducing the infection rate.
 We compute $100$ different time-series for $\beta$ by assuming that the vector of NPIs, $\mathbf{x}^t$, randomly changes each week. For each of the resulting time series, we generate $20$ data points equally spaced from day $0$ to day $140$ of the outbreak evolution. The generation process relies on a highly accurate Euler integration with $10.000$ iterations per time step and uses the same initial condition and $\gamma$ value described in the \emph{Training Procedure} experiment (\Cref{sec:training_procedure}).
To approximate the non-linear dependence in \Cref{eq:beta_series}, we rely on a DNN which forecasts the value of $\beta$ based on $\mathbf{x}^t$ and the state of the system at time $t-1$. Thus, the resulting universal approximator of the \uodeacronym{} framework is a black-box model able to capture complex functional dependencies, but lacking interpretability. 

\begin{figure}[htp]
   \centering
   \includegraphics[width=1\linewidth]
   {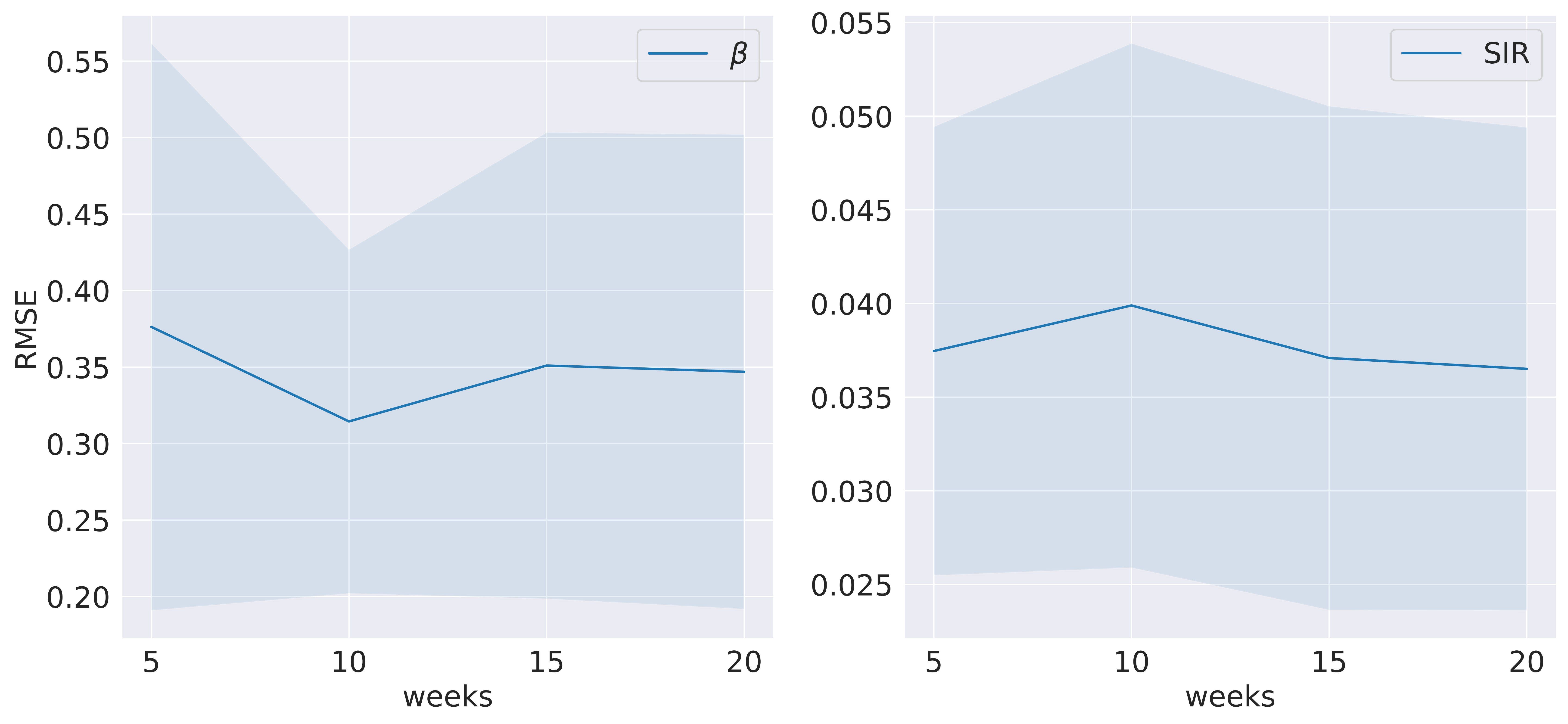}
 \caption{Non-Linear dependence: $\beta$ and prediction errors with different sampling frequencies.} 
 \label{fig:exp-nl}
 \end{figure}

The experimental results show that the \uodeacronym{} framework is able to estimate the dynamic system state with high accuracy (the RMSE of the state prediction is  $0.037 \pm 0.013$); however, the model is unable to provide an accurate estimation of the $\beta$ time-series, which RMSE is equal to $0.37 \pm 0.17$.
Similarly to the RC circuit, we investigate the effect of the data sampling frequency to the parameter approximation accuracy of the \uodeacronym{}. 
We consider $4$ different time horizons, namely $5, 10, 15$, and $20$ weeks of the outbreak evolution, in which we sample $20$  equally spaced data points. 
We train the model on the resulting data, and 
we compute the reconstruction error (RMSE) on the complete epidemic curve of $20$ weeks. We report both the parameter approximation error and the curve reconstruction error in Figure \ref{fig:exp-nl}.

Conversely to RC circuit, the sampling process does not seem to a have a significant impact on the model accuracy. The reason for this result may be found in the complexity of the function to be approximated, and in the impact of $\beta$ parameter to the epidemic curve. In the RC-circuit, the system state evolution is an exponential function of the unknown parameter $\tau$, and we can design the collection process to cover the temporal interval where the impact of $\tau$ is more relevant. In the SIR model, we do not have a closed-form of the epidemic curve, and thus it is harder to select the most relevant temporal horizon.

\section{Conclusions}

\label{sec:conclusions}

In this paper, we perform an in-depth analysis of the UDEs framework in solving data-driven discovery of ODEs. We experimentally probe that \pairmethod{} gradient descent is faster than the \unrollmethod{} version without compromising the final performances. We highlight some issues arising when combining data-driven approaches and numerical integration methods, like the discrepancy in accuracy between state evolution prediction and system parameter approximations. We investigated the integration method precision as a possible source of error, and we discuss the trade-off between approximation accuracy and computational time. Moreover, we study the importance of the data collection process in reaching higher parameter approximation accuracy.

We believe that our analysis can foster the scientific community to further investigate the capabilities and limitations of Physics-informed machine learning frameworks in the context of differential equation discovery. 

In the future we plan to extend our analysis by i) testing different numerical integration solvers (e.g., higher-order Runge-Kutta or adjoint-state methods), ii) considering the unknown parameters to be stochastic, rather than deterministic, iii) extending the analysis to PDEs.

%
%
%
\bibliographystyle{unsrt}  
\bibliography{references}

\end{document}